\documentclass[a4paper, 10pt, conference]{ieeeconf}      
\usepackage{FG2026}
\usepackage{booktabs}
\usepackage{multirow}
\usepackage[table]{xcolor}
\usepackage{adjustbox}
\usepackage{subcaption}

\FGfinalcopy 

\IEEEoverridecommandlockouts                              
\usepackage{graphics} 
\usepackage{graphicx}
\usepackage{epsfig} 
\usepackage{times} 
\usepackage{amsmath} 
\usepackage{amssymb}  
\definecolor{cinza}{cmyk}{0,0,0,0.5}

\def\FGPaperID{324} 

\title{\LARGE \bf
Revisiting Emotions Representation for Recognition in the Wild
}

\author{\parbox{16cm}{\centering
    {\large Joao Baptista Cardia Neto$^1$, Claudio Ferrari$^2$ and Stefano Berretti$^2$}\\
    {\normalsize
    $^1$ São Paulo State Technological College (FATEC), São Paulo, Brazil\\
    $^2$ Media Integration and Communication Center (MICC) \\ Department of Information Engineering, University of Florence, Italy}}
}

\begin{document}

\ifFGfinal
\thispagestyle{empty}
\pagestyle{empty}
\else
\author{Anonymous FG2026 submission\\ Paper ID \FGPaperID \\}
\pagestyle{plain}
\fi
\maketitle

\begin{abstract}
Facial emotion recognition 
has been typically cast as a single-label classification problem of one out of six prototypical emotions. However, that is an oversimplification that is unsuitable for representing the multifaceted spectrum of spontaneous emotional states, which are most often the result of a combination of multiple emotions contributing at different intensities. Building on this, a promising direction that was explored recently is to cast emotion recognition as a distribution learning problem. Still, such approaches are limited in that research datasets are typically annotated with a single emotion class. In this paper, we contribute a novel approach to describe complex emotional states as probability distributions over a set of emotion classes. To do so, we propose a solution to automatically re-label existing datasets by exploiting the result of a study in which a large set of both basic and compound emotions is mapped to probability distributions in the Valence-Arousal-Dominance (VAD) space. In this way, given a face image annotated with VAD values, we can estimate the likelihood of it belonging to each of the distributions, so that emotional states can be described as a mixture of emotions, enriching their description, while also accounting for the ambiguous nature of their perception. In a preliminary set of experiments, we illustrate the advantages of this solution and a new possible direction of investigation. Data annotations are available at https://github.com/jbcnrlz/affectnet-b-annotation.
\end{abstract}

\section{Introduction}\label{sec:intro}
Automatically determining the emotional state of a person by observing his/her facial expression is a long-standing problem in computer vision. Addressing this task requires solving two difficult sub-problems: \textit{(i)} how to effectively represent and describe emotional states, and \textit{(ii)} how to get annotated data for training machine learning algorithms. 

\begin{figure}[!t]
    \centering
    \includegraphics[width=0.99\linewidth]{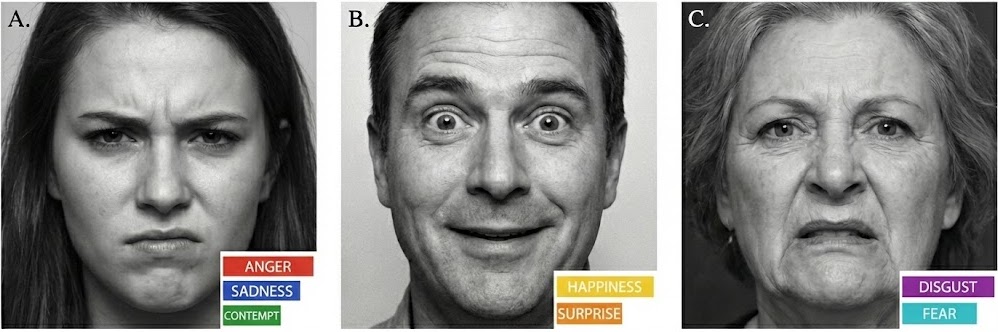}
    \caption{Example of the intrinsic ambiguity of emotion display and perception. The three faces portray expressions that clearly reveal a combination of different emotions contributing at different intensities.}
    \label{fig:emo_ambig}
\end{figure}

Since the initial stages of exploration, the problem has been cast as that of categorizing a facial image into a predefined set of prototypical, universally recognized emotion categories, \textit{i.e.}, neutral, angry, disgust, fear, happy, sad, surprise, and, more rarely, contempt~\cite{ekmann1973universal}. 
However, such a protocol represents an oversimplification, as the real spectrum of genuine emotional states is significantly multifaceted. According to Plutchik's theory~\cite{plutchik1980general} and previous works~\cite{kawamura2025enhancing,zhou2015emotion}, genuine emotional states are actually a subtle mixture of emotions that exist in varying degrees of intensity, which suggests that the single-label representation of emotions is just an inadequate approximation (Figure~\ref{fig:emo_ambig} showcases a clear example). Attempts have been made to comply with this interpretation by introducing the concept of \textit{compound emotions}~\cite{du2014compound}, in which a finite set of composed classes is defined, \textit{e.g.}, happily surprised, angrily disgusted. Despite being preferable, it appears restrictive in that it still requires the definition of some predefined categories.
Recently, a promising direction of investigation was found in casting emotion recognition as a label distribution learning problem~\cite{jia2019facial, kawamura2025enhancing, le2023uncertainty, zhou2015emotion}. Such approaches formalize the problem as a multi-label classification, in an attempt of relaxing the hard-label constraint and so account for the intrinsic ambiguity of emotions.
To go beyond hard-categorization and provide a continuous and natural description of emotions, the concept of \textit{dimensional emotions}~\cite{schlosberg1954three} was introduced, which describes emotions in terms of three continuous values, namely \textit{Valence} (pleasant/unpleasant), \textit{Arousal} (activated/calm) and \textit{Dominance} (control/submissive), or VAD. With such, any complex emotional state can potentially be described. Yet, they are of difficult interpretation and require trained experts, making it difficult to link VAD values to common and clearly understandable emotional states~\cite{guerdelli2023towards}. 

Regardless of the setting, developing a recognition method requires annotated data. Initially, datasets were collected in lab-controlled conditions, where subjects were instructed to replicate specific expressions, hence avoiding the annotation process. The reason to do so is that humans display and perceive emotions subjectively, making data labeling a difficult and ambiguous process~\cite{kawamura2025enhancing}. The latter quickly became obsolete as soon as the problem naturally shifted to more realistic scenarios, where subjects act spontaneously and thus cannot be instructed to replicate specific expressions. The typical solution that was found for labeling, as people with different backgrounds might interpret facial expressions differently, is that of involving multiple users and labeling the data in a majority-voting fashion. However, this usually leads to uncertainty, disagreement, and thus ambiguities in the annotation~\cite{guerdelli2023towards,kawamura2025enhancing, le2023uncertainty}. Such nuisances arise as a result of forcing a hard categorization of emotions. 

This paper revolves around the idea that current protocols for emotion recognition are flawed. Trying to assign specific categories to emotions, which are naturally multifaceted, appears conceptually wrong. It would be as choosing between ``green'' and ``blue'' to label a cyan object; \textit{cyan is neither of them, it is both}. Consequently, even estimating the accuracy of methods in turn becomes flawed. Building on this idea, in this paper, we investigate a new exploratory solution to describe emotional states as probability distributions over a set of basic emotion classes, and a procedure to automatically re-label existing datasets with such. We aim at exploiting the continuous nature of the valence-arousal-dominance description of emotions, and propose a solution for linking such numerical values with a mixture of categorical emotion terms. Specifically, we exploit the findings of Russel \textit{et al.}~\cite{RUSSELL1977273}, who defined a mapping between $151$ human-readable emotion terms, \textit{e.g.}, happiness, sadness, rage, anxiety, joy, etc., to probability distributions over the VAD space. We exploit such mapping to build a probabilistic mixture model of emotions in the VAD space so that, given a face image annotated with VAD values, we can use it to estimate the likelihood of the image to belong to each emotion category, \textit{i.e.}, component of the mixture, ultimately obtaining a fine-grained label that richly describes the emotional state as a weighted mixture of multiple emotions. Finally, we use the re-labeled data to design and train an architecture for estimating complex emotional states. In sum, the contributions of this paper are:
\begin{itemize}
    \item We propose an alternative paradigm for emotion recognition in which emotional states are defined and predicted as probability distributions of a set of emotions;
    \item We propose an approach that, given VAD values and emotion labels, estimates a probability distribution of emotions, allowing us to automatically re-label existing datasets without further manual intervention;
    \item We validate this approach by designing a user interface for annotating face images in a multi-label fashion. It outputs a probability distribution describing how much a predefined set of emotions contributes to the overall emotional state perception;
    \item We develop a baseline architecture that uses the re-labeled dataset to learn predicting emotional states as probability distributions, and propose alternative metrics for performance assessment.
\end{itemize}

\section{Related Work}
\label{sec:related}
In a broad sense, it is possible to divide the vast literature on automatic facial emotion recognition (FER) into three main categories. 

The first category, includes methods categorizing a facial expression into one of the six discrete universal emotion categories based on Ekman's work~\cite{ekmann1973universal}. This approach is the most used, being easy to interpret and intuitive to humans~\cite{farzaneh2021facial, yang2018facial, zeng2018facial, zhang2022learn}. 

Methods in the second category estimate continuous Valence and Arousal values, also known as dimensional emotion recognition~\cite{deng2020mimamo, liang2025mamba, meng2022valence,
mitenkova2019valence, yu2024improving}; 

The third category comprises uncertainty-based label distribution learning (LDL)~\cite{gao2017deep} methods, which exploit the uncertainty inherent in the task~\cite{kawamura2025enhancing,matin2020hey}. They ground on the observation that the perception of emotions can change significantly from individual to individual, resulting in noisy and often inconsistent labels~\cite{guerdelli2023towards,kawamura2025enhancing}. 
In the literature, solutions have been proposed that either try to fuse multiple datasets and generate pseudo labels or, similarly, use soft-label augmentation~\cite{kawamura2025enhancing}, or even use auxiliary networks to estimate the uncertainty~\cite{matin2020hey}. 
In~\cite{jia2019facial}, the annotation ambiguity is exploited to learn correlations across emotions and facial features to improve prediction. The study is conducted on the s-JAFFE lab-controlled dataset,  which extends the JAFFE database~\cite{lyons1998coding} by including multi-user annotation per image, so that the disagreement can be used as an uncertainty clue. More recently, Le \textit{et al.}~\cite{le2023uncertainty} proposed a way to extend the LDL approach to standard datasets. They use VA values to compute proximity measures across images and build an uncertainty model to estimate the emotion of a sample face from observing its neighbors in the VA plane. The method is promising, yet it still produces a single-class prediction and is limited to the six basic emotion classes. 
In contrast, in this work, we differentiate with respect to the literature by proposing a new method for learning a mixture model of emotions. In doing so, given a face image annotated with VAD values, we can generate a probability distribution of emotions so that its description is enriched.


\section{Methodology} \label{sec:method}
To automatically re-label face images with probability distributions of emotions, we use the mapping proposed by Russel~\cite{RUSSELL1977273}. This mapping associates an emotion term $E$ (say, ``happiness'') to a normal distribution in the VAD space, that is $E \rightarrow e \sim \mathcal{N}(\boldsymbol{\mu}_e,\boldsymbol{\sigma}_e)$, where $\boldsymbol{\mu}_e = (\mu_V, \mu_A, \mu_D)$ and $\boldsymbol{\sigma}_e = (\sigma_V, \sigma_A, \sigma_D)$, the subscripts indicating Valence, Arousal and Dominance, respectively. Figure~\ref{fig:va_full} gives an example of the distributions for the 6 universal emotions. The reader can appreciate how some of them have a large overlap, revealing how emotional states cannot be clearly separated, suggesting that different emotions are indeed entangled and fluid, and that their representation requires taking this ambiguity into account. Our approach assumes that VAD annotations are available.


\subsection{Recovering dominance dimension}\label{subsec:dominance-est}
\begin{figure}[!thb]
    \centering
    \includegraphics[width=0.85\linewidth]{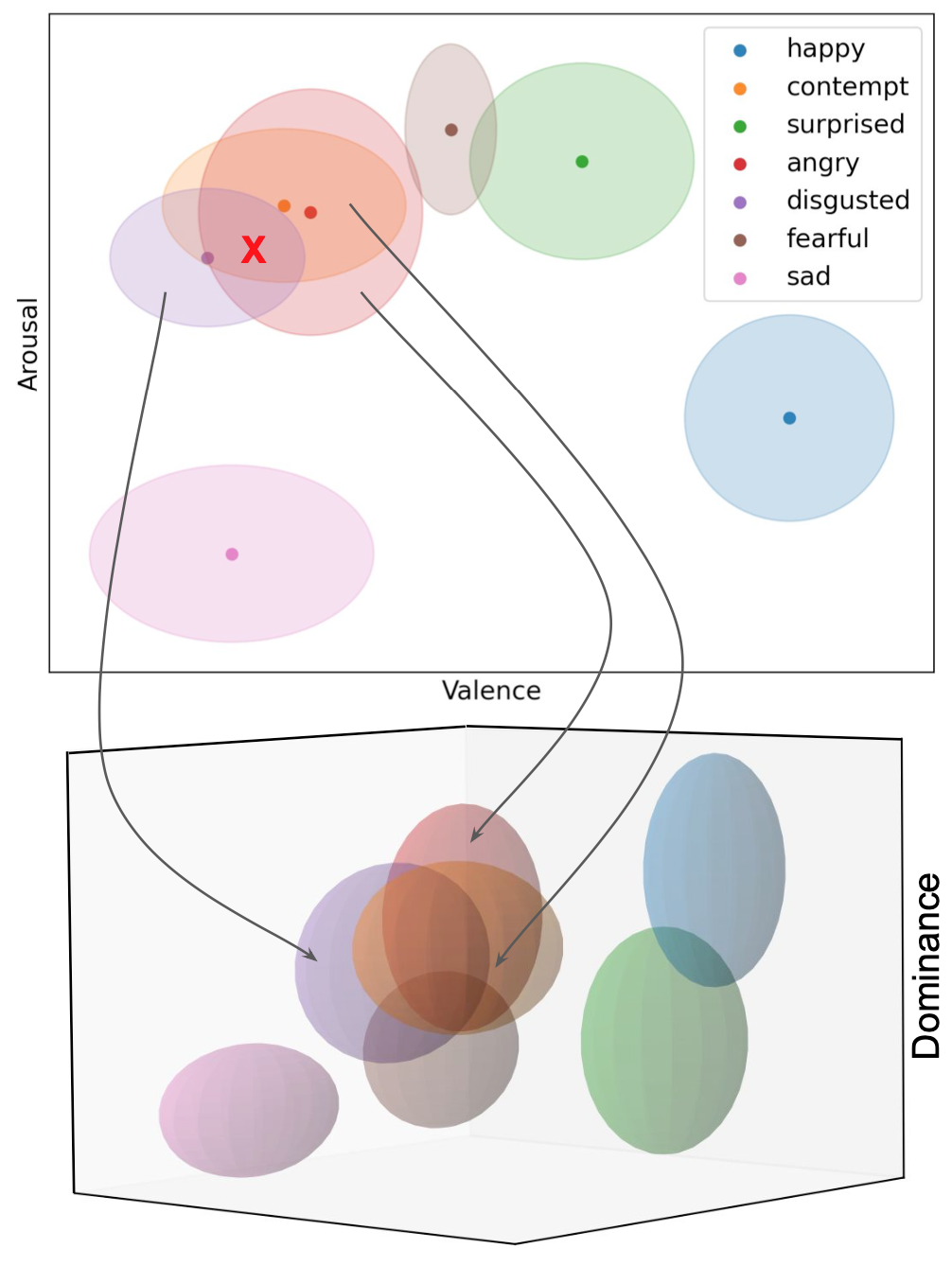} 
    \caption{2D/3D visualization of the valence and arousal space of the 7 universal emotions. Some of them severely overlap in the VA plane, thus making the mapping of a point (\textit{e.g.}, the \textcolor{red}{X}) to one or more emotion labels ambiguous. Adding the dominance dimension (bottom graph) helps disambiguate between them and improve the mapping.}
    \label{fig:va_full}
\end{figure}

In the current literature, all existing datasets annotated with valence-arousal values do not include the dominance dimension, yet it is imperative to add it into the process. Indeed, Figure~\ref{fig:va_full} shows that some emotions share similar valence-arousal values and thus overlap in the VA plane, \textit{e.g.}, ``disgust'' and ``angry'' but can be better distinguished via dominance. Thus, given the VAD distributions defined by~\cite{RUSSELL1977273}, we try to recover the dominance value from the valence-arousal pair and add it to the data annotation. 
%
%

However, it would be ambiguous to estimate the dominance for emotions that overlap in the VA plane without additional clues. So, we employ the information from the categorical labels to disambiguate. Specifically, we utilize a Combined Weighted Dominance Estimation (CWDE), which combines a standard linear regression with specific correlation models for each of the universal emotions $\mathcal{E}^u=\{E_i\}_{i=1:6}$. Using the prior given by the label $E_i$ serves to condition the estimation of the dominance value in the context of the most likely emotion. Given a VA pair ($V_{obs}$, $A_{obs}$), we calculate the likelihood of it belonging to each universal emotion $E_i$:
\begin{equation}
\label{eq:cwd_like}
    P(V_{\text{obs}}, A_{\text{obs}} | \text{E}_i) \propto P(V_{\text{obs}} | \text{E}_i) \times P(A_{\text{obs}} | \text{E}_i) .
\end{equation}    

\noindent
Note that we assume that V and A are independent, univariate Gaussian distributions. The calculated probability is normalized to obtain the posterior, representing the plausibility that the observed valence-arousal belongs to the emotion $E_i$, and we use it as a weight for estimating dominance:
\begin{equation}
    w_i = P(\text{E}_i | V_{\text{obs}}, A_{\text{obs}}) \propto P(V_{\text{obs}}, A_{\text{obs}} | \text{E}_i) \times P(\text{E}_i) .
\end{equation}
We assume that the prior is equiprobable for all the emotions. The final Dominance ($\hat{D}$) is calculated as the weighted mean of the likelihood for each of the regression models ($\mathcal{F}_i$) for the emotions:
\begin{equation}
    \hat{D} = \sum_{i \in \mathcal{E}^u} w_i \times \mathcal{F}_i(V_{\text{obs}}, A_{\text{obs}}) . 
\end{equation}

For defining $\mathcal{F}_i$, we assume that all the variables are normally distributed, and utilize the following equation: 
%
\begin{equation}
    \mathcal{F}_i = \beta_0 + \beta_1 V_{\text{obs}} + \beta_2 A_{\text{obs}} .
\end{equation}
Both $\beta_1$ and $\beta_2$ are calculated from the first and second order statistics of a specific emotion (mean $\mu$, standard deviation $\sigma$, and Pearson coefficient $\rho$). For each of the universal emotional states, we use closed-form solutions, considering cross-correlation between predictors:
\begin{equation}
    \beta_1 = \frac{\sigma_D}{\sigma_V} \left( \frac{\rho_{VD} - \rho_{VA} \rho_{AD}}{1 - \rho_{VA}^2} \right) ,
\end{equation}
\begin{equation}
    \beta_2 = \frac{\sigma_D}{\sigma_A} \left( \frac{\rho_{AD} - \rho_{VA} \rho_{VD}}{1 - \rho_{VA}^2} \right) ,
\end{equation}
\begin{equation}
    \beta_0 = \mu_D - \beta_1 \mu_V - \beta_2 \mu_A .
\end{equation}
In these equations, the subscripts $D,V,A$ indicate dominance, valence, and arousal, respectively. At the end of the process, the VA annotation of face images is integrated with a plausible dominance value. Note that this is not going to be used during training or testing, but serves as a means to build a more accurate representation of emotions.

\subsection{Terms selection}\label{subsec:term-select}  
In~\cite{RUSSELL1977273}, Russell defined 151 emotional states, some of which are extremely similar and redundant, \textit{e.g.}, angry and upset. On the one hand, manually selecting a subset of them would be arbitrary, posing concerns about the fairness of the selection. On the other hand, using all of them would be problematic if not enough data are available for each possible class. To strike a balance between description granularity and feasibility, we join similar emotions by using their intersection ratio in the VAD space as a criterion. Given an emotional distribution $e_i \sim \mathcal{N}(\boldsymbol{\mu}_e,\boldsymbol{\sigma}_e)$, we utilize a KD-tree based on the mean VAD values to select 5 next-neighbors, and use the MonteCarlo method for estimating the intersection volume between each pair of distributions $e_i, e_j$, defined as: 
\begin{equation}
\label{eq:volIntersec}
    V_{\cap} = \int_{\mathbb{R}^3} \min(\text{PDF}_i(V, A, D), \text{PDF}_j(V, A, D)) \, dV \, dA \, dD .
\end{equation}
The final value is normalized by the volume of the smaller one, resulting in the normalized intersection measure (NIM):
\begin{equation}
\label{eq:nim}
    \text{NIM}_{i,j} = \frac{V_{\cap}}{\min(V_i, V_j)} . 
\end{equation}

We define that the universal emotions, plus neutral, would be utilized, without any type of fusion. For the rest of the emotional states, we fuse the ones having NIM above some threshold $t$. Changing $t$ allows us to control the granularity of description. We empirically set $t=0.5$, indicating that more than 50\% of the areas overlap. In such cases, we remove both the old distributions from the taxonomy and add the new fused one. To fuse two emotion distributions and account for their specific means and standard deviations, we generate 1000 random points from each of the original distributions and calculate the mean and standard deviation of the junction of the generated points for each of the distributions. 
The process is repeated iteratively until there are no more emotions that have a $NIM > t$.

By controlling the value of $t$, one can create sets of emotions of different numerosity. Using a lower $t$ value would result in more emotion classes to be generated, at the cost of having greater ambiguity, \textit{i.e.}, overlap, between them. Greater $t$ values lead to fewer emotion classes but with a larger extent in the VAD space, thus covering more of the underlying classes that generated them.

\subsection{Estimating the likelihood} \label{subsec:likelihood-est}
To project the VAD values to the labels in our fused taxonomy, we use a likelihood-based mapping. Given VAD values, we estimate the likelihood of that point belonging to one of the $K$ emotion classes. The likelihood is calculated as the sum of the product of the PDF of the univariate distributions for $V$, $A$, and $D$ conditioned on the emotion $E_k$. We use a logarithmic approach to ensure numerical stability: 
\begin{equation}
\label{eq:likelihood}
\begin{split}
    \log(P(\mathbf{x} | \mathcal{E}_k)) 
    =  -\frac{1}{2}\left(\frac{x - \mu_k}{\sigma_k}\right)^2 - \log(\sigma_k \sqrt{2\pi}) ,
\end{split}
\end{equation}
\begin{equation}
\begin{split}
    \log(L_k) = \log(P(\mathbf{x} | \mathcal{E}_k)) = \log(P(V|\mathcal{E}_k)) +  \\ + \log(P(A|\mathcal{E}_k)) + \log(P(D|\mathcal{E}_k)) .
\end{split}
\end{equation}

%

As the final step for this process, we use $L_k$ to calculate the posteriori in which the class $\mathcal{C}_k$ is the true label. 
Assuming that classes are all equiprobable, the soft label $p_k$ for a class $\mathcal{C}_k$ is given by the normalization of the likelihood:
\begin{equation}    
p_k = P(\mathcal{C}_k | \mathbf{x}) \propto \frac{P(\mathbf{x} | \mathcal{C}_k)}{\sum_{j=1}^{K} P(\mathbf{x} | \mathcal{C}_j)} .
\end{equation}
In the logarithm space, we use the function LogSumExp (LSE) to calculate the normalization term:
\begin{equation}
\log(p_k) = \log(P(\mathbf{x} | \mathcal{C}_k)) - \text{LSE}_{j=1}^{K} \left(\log(P(\mathbf{x} | \mathcal{C}_j))\right) .
\end{equation}

At the end of the process, we obtain  $p = [p_1, p_2, \dots, p_K]$ with $\sum p_k = 1$. The $p$ vector is the new probabilistic ground-truth description of the image.

\subsection{Dataset Re-Balance }  
\label{subsec:data_balance}
\begin{figure}[!t]
    \centering
    \includegraphics[width=0.98\linewidth]{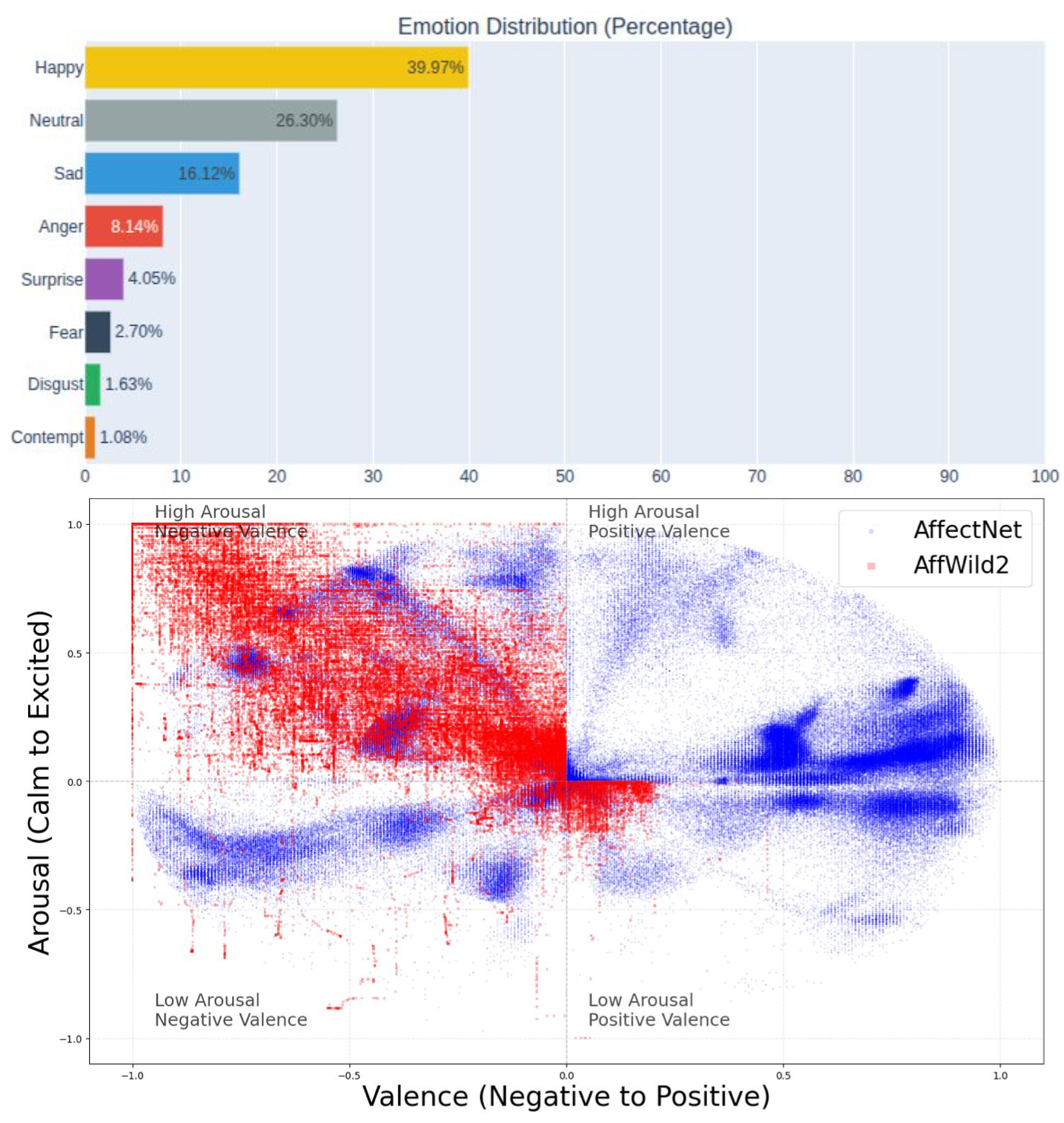}
    \caption{Emotion distribution for the AffectNet dataset. {\bf Top}: there is a clear imbalance towards ``happy'' label, while negatives have way less examples. {\bf Bottom}: We added images from AffWild2 (red points) choosing them on the 4th quadrant of the VA plane so to balance negative emotions.}
    \label{fig:affecnet_distro}
\end{figure}

The goal of the proposed approach is to exploit emotion recognition datasets that come with valence-arousal annotations and re-label them with the estimated emotion distributions so that novel learning algorithms can be designed. Among all the existing datasets in the literature, we opted to utilize the AffectNet~\cite{mollahosseini2017affectnet} dataset, given its widespread use in the community, and being annotated with both valence-arousal and emotion labels. However, given its unbalanced nature (Figure~\ref{fig:affecnet_distro}, top), we tried to compensate for the impact of emotions that were less represented. To do so, we observed the VA space instead of the quantity of emotion labels, and we tried to equalize the sample density on the VA space. The reason is that our approach uses the VA values to estimate the probability for each emotion; thus, with a view to improve the estimation, we prioritized an equal proportion of samples populating the VA plane rather than the frequency of the emotion categories.

We utilize the training data from the AffWild2 dataset~\cite{kollias2019deep} as additional data. To equalize the sample density, we use a simple density ratio criterion: for each quadrant of the VA plane, its density is defined as the number of examples in the quadrant divided by the total area of the quadrant (Figure~\ref{fig:affecnet_distro}, bottom). We stipulated that the quadrant density of happy emotion, which is the most frequent, would be the maximum allowed quadrant density. Given a face from AffWild2, we utilize its valence and arousal value to determine which quadrant that image would fall into. If it falls in a quadrant where the density is less than the maximum quadrant density, the image is added to the dataset, and the density of that quadrant is recalculated. We utilized the valence and arousal values, rather than the emotion annotation, because we used those values to generate the new ranking annotation. In this sense, if we get values into a low-density region and fill it with more examples, even if it is not the class that has fewer images, it would generate a bigger diversity in the dataset. A total of $59,845$ images from AffWild2 have been added to the dataset. We will refer to this extended set as ``\textit{B-AffectNet}''. 

\section{Data Validation} 
\label{sec:annot_val}
After the re-labeling process is completed, all the B-AffectNet training and testing images are labeled with emotion distributions. Before proceeding, it is important to measure the validity of the annotations. However, doing so quantitatively is complicated as there is no ground-truth. One possible option would be to take the predominant emotion from the distributions and compare it to the ground-truth label. However, we noted that in many cases, the VA annotations and the label are inconsistent, which would make the evaluation biased. Inconsistencies also occur among different annotators as well (in~\cite{mollahosseini2017affectnet} it is reported that the average level of label agreement between annotators is around 60\%). Thus, after careful consideration, we opted for collecting distribution annotations from users, specifically instructed to label a subset of the data with annotations that ground on the proposed distribution-based representation, and finally compare those with our automatic result.

To this aim, we developed a web interface where users are shown a face image and can select how much each basic emotion is perceived using a slider. We opted for using the 7 universal emotions in this test. The user can provide, independently for each emotion, a score $s_e \in [0,1]$ determining how much they perceive a specific emotion on the displayed face. The sliders range from 0 (the emotion is not perceived at all) to 1 (the emotion is very clearly expressed). We did not include ``\textit{neutral}'' among the user choices, as we deem it could create confusion. In fact, a neutral emotional state can be interpreted either as the absence of any emotion or a degree of intensity when expressions are subtle. Given our formulation of the emotion description, we opted for taking the second interpretation. Thus, users are clearly instructed not necessarily indicate the presence of emotions if they do not perceive any, and we recover ``neutral'' from the rest. 

\begin{figure}[!t]
    \centering
    \includegraphics[width=0.95\linewidth]{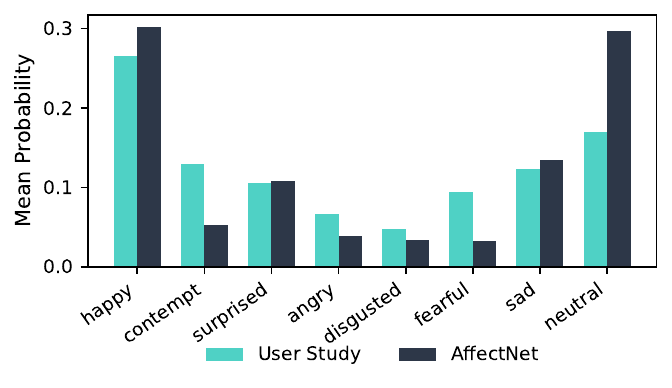} 
    \caption{Comparison between the average likelihood across all samples of the user study and our automatic approach.}
    \label{fig:studies_comparison}
\end{figure}

After the user annotated an image, we obtain a score $s_e$ for each emotion, we compute the value for ``neutral'' and normalize the values. In particular, there are two cases:
\begin{itemize}
    \item The sum of the scores is $\sum_{e=1}^{7} s_{e} \ge 1$. In this case, we normalize the values such that the sum adds to 1 and can be interpreted as a probability distribution. Neutral is given the value 0;
    \item The sum of the scores is $\sum_{e=1}^{7} s_{e} < 1$. In this case, we assign the score of the neutral state as $s_n =1-\sum_{e=1}^{7} s_{e}$.
\end{itemize}

For each session, the user receives a random quantity of images, from one to four. The user can take as many sessions as he/she wants, but an image can only be annotated once. A specific image is no longer shown to any user after it received at least 3 annotations.

\subsection{Data Validation Results} \label{subsec:data_val_res}
We collected annotations from 22 participants, who annotated a total of 126 images. These are used to compare those from our automatic process. As Figure~\ref{fig:studies_comparison} illustrates, the probability of the emotions in the user study and on our estimated annotation is very similar. This is a strong evidence that our process is close to the perception of emotions in the real world. To measure the discrepancy, we calculated the Jensen-Shannon (JS) and Kullback-Leibler (KL) divergence, reporting a mean value of $0.21$ and $1.75$.

The emotions with the biggest discrepancy are contempt, fearful and neutral. In the case of contempt and fear, they are more prevalent in the user study than in our annotation. A reason could be that, in the VAD space, the distributions of those emotions lie in an area in which there is heavy overlap between other negative strong emotions, more specifically disgust and anger (see Figure~\ref{fig:va_full}). This makes it hard to clearly discriminate them, and indeed the automatic process tends to balance their contribution. This could appear as an artificial bias, yet the user study reports lower probability for them as well, meaning that such emotions are likely difficult to identify even for humans. Neutral, on the other hand, is more prevalent in the automatic annotation than in the user study. This can be a result of how neutral is defined in our scenario; users do not point neutral directly, but is recovered a posteriori. We argue that this is a clear hint that complete neutrality is not really a perceivable state, and that in spontaneous conditions emotions are perceived even if at very low intensity, making the use of ``neutral'' as a specific emotion class quite improper for the task.  

\begin{figure}[!t]
    \centering
    \includegraphics[width=0.99\linewidth]{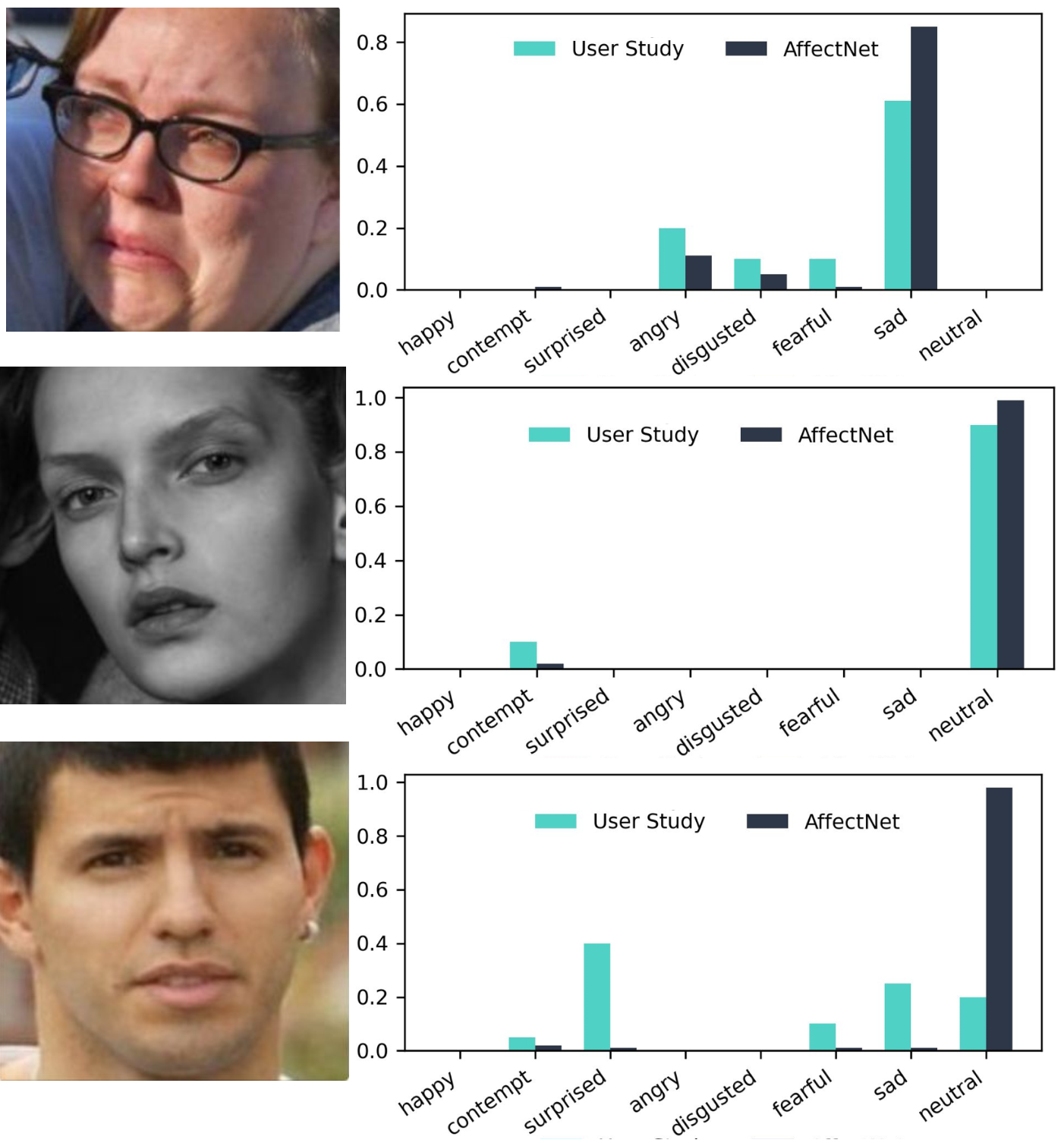}
    \caption{Some example comparison between our automatic annotations (AffectNet) and the values of the user study.}
    \label{fig:faces_expr}
\end{figure}

Figure~\ref{fig:faces_expr} shows an example of the result of our annotation compared to users annotations for a few specific faces. In Fig.~\ref{fig:faces_expr} top and middle rows, there is a quite high level of agreement between what we estimated and what the user has annotated, while in the bottom row the discordance is higher. Users perceived a mixture of emotions rather than clear neutrality. The latter fact again suggests that the interpretation of the ``neutral'' class is ambiguous when expressions are subtle, since neutral is indicated towards being an absence of emotions. Further investigations are required on this matter as it is also possible that users in this scenario were biased towards putting an emotion in which they think is more prevalent. Nevertheless, given how subjective and hard the task is, these results support the feasibility of our proposed strategy for enriching the annotation of emotions, highlighting their similarity with natural human perception. 

\section{Model Architecture} 
\label{sec:model_arch}
\begin{figure}[!h]
    \centering
    \includegraphics[width=0.99\linewidth]{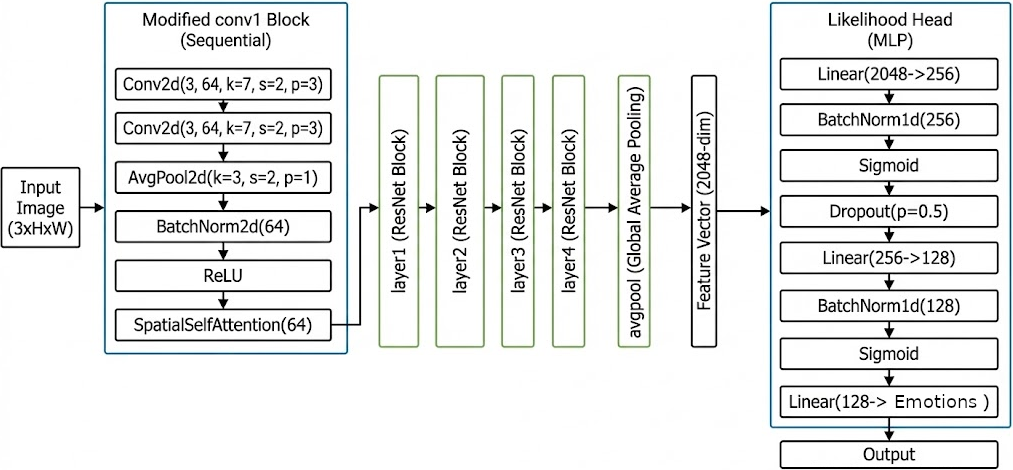} 
    \caption{The structure of the proposed baseline model.}
    \label{fig:archNet}
\end{figure}

To train and predict the emotional state of a face as a probability distribution, we designed a simple baseline network architecture. This allows us to assess the practical applicability of our proposal in a learning framework.
Our model first processes the image via convolutional layers and utilizes a self-attention to focus on areas of the face that have more discriminative information towards emotions. The output is fed to a ResNet50 backbone, whose output goes through a Likelihood head. The latter acts as a non-linear mapping that projects the feature vector into a $C$-dimensional space, being $C$ the number of desired emotions. Figure~\ref{fig:archNet} illustrates the full model. 

\begin{table*}[ht!]
\centering
\caption{Results comparing different loss functions on the 8 emotion setting. Best results in bold, second-best underlined.}
\label{tab:loss_comparison}
\begin{adjustbox}{width=\textwidth}
\begin{tabular}{lcccccccc}
\toprule
\textbf{Experiment} &
\textbf{Acc. (\%)} &
\textbf{F1-Score} &
\textbf{Precision} &
\textbf{Recall} &
\textbf{JS ($\downarrow$)} &
\textbf{KL ($\downarrow$)} &
\textbf{Cosine ($\uparrow$)} &
\textbf{Pearson ($\uparrow$)} \\
\midrule

SCN~\cite{SCN} &
60.23 &
 - &
 - &
 - &
 - &
 - &
 - &
 - \\ 

PSR~\cite{PSR} &
60.68 &
 - &
 - &
 - &
 - &
 - &
 - &
 - \\ 

DAN~\cite{DAN} &
\underline{62.09} &
 - &
 - &
 - &
 - &
 - &
 - &
 - \\

EfficientFace~\cite{effFace} &
60.23 &
 - &
 - &
 - &
 - &
 - &
 - &
 - \\

MA-Net~\cite{MANET} &
60.29 &
 - &
 - &
 - &
 - &
 - &
 - &
 - \\

POSTER++~\cite{POSTER++} &
\textbf{63.77} &
 - &
 - &
 - &
 - &
 - &
 - &
 - \\

\midrule
 
FocalLoss &
30.68 &
0.22 &
0.21 &
0.31 &
0.27 &
3.63 &
0.57 &
0.13 \\

ConsistencyLoss &
42.11 &
\textbf{0.37} &
0.35 &
\textbf{0.42} &
\textbf{0.21} &
\textbf{1.99} &
\textbf{0.65} &
\textbf{0.42} \\


GuidedConsistencyLoss &
37.41 &
0.28 &
0.26 &
0.37 &
0.26 &
\underline{3.77} &
0.57 &
0.20 \\

RegularizedConsistencyLoss &
39.78 &
0.31 &
0.30 &
0.40 &
\underline{0.23} &
4.08 &
\underline{0.59} &
0.27 \\

\midrule

GuidedConsistencyLoss -- LD &
41.06 &
0.34 &
\textbf{0.46} &
\underline{0.41} &
0.33 &
6.01 &
0.54 &
0.35 \\

RegularizedConsistencyLoss -- LD &
42.74 &
\underline{0.36} &
\underline{0.36} &
\textbf{0.42} &
0.31 &
5.28 &
0.52 &
\underline{0.38} \\
\bottomrule
\end{tabular}
\end{adjustbox}
\end{table*}

\subsection{Emotion Consistency Loss}
The dataset we are working with is unbalanced. Happy and neutral faces are prevalent over other types of emotions. With this in mind, we initially opted to train models with the Focal Loss~\cite{lin2017focal}, which is a modified cross-entropy that addresses class imbalance by dynamically down-weighting easy (common) examples and focusing training on hard (rare) examples.  After initial experiments, we soon found out that the data unbalance was not the only problem. We observed that in the VAD space some distributions overlap but we should take into consideration the semantic consistency between emotions. For instance, someone can be simultaneously happy and surprised, but cannot be happy and sad at the same time~\cite{russell1980circumplex}. This has motivated us to propose the \textit{Emotion Consistency Loss} and some variations of it. This loss utilizes a conflict matrix that captures the mutual exclusivity relations between pairs of emotions. The idea is that the model learns this structure during training time, and it penalizes pairs that are conflicted (e.g., happy and sad). To ensure that the learning is guided by biological and psychological principles, we have utilized the sources~\cite{plutchik1980general,russell1980circumplex} as prior knowledge. Basically, the term penalizes the joint activation of conflicting classes as follows:
\begin{equation}
\mathcal{L}_{cons} = \sum_{i}^{n} \sum_{j > i}^{n} \sigma(W_{i,j}) \cdot (p_i \cdot p_j) ,
\end{equation}
with $W$ being the conflict matrix, $n$ the number of emotions, $\sigma$ a sigmoid function, $p_i$ the probability of the emotion represented in the face being the emotion $i$, and $p_j$ the probability of being emotion $j$.

To avoid the conflict matrix becoming highly dense, which could lead to sub-optimization of the probabilities from the classes, we apply an L1 regularization on the matrix parameters. The idea is to ensure sparsity, making sure that only statistically significant conflicts are kept. The sparsity regularization is given by:
\begin{equation}
    \mathcal{L}_{sparse} = \frac{1}{N^2} \sum_{i} \sum_{j} |W_{i,j}| .
\end{equation}
The full loss function is obtained as:
\begin{equation}
    \mathcal{L}_{total} = \mathcal{L}_{FL} + \mathcal{L}_{cons} + \lambda \mathcal{L}_{sparse} ,
\end{equation}
with $\mathcal{L}_{FL}$ being the Focal Loss, $\mathcal{L}_{cons}$ being the consistency term with the conflict matrix, $\mathcal{L}_{sparse}$ the sparsity regularization term, and $\lambda$ is a hyperparameter for the weight of the sparsity regularization.

We carried out experiments with three variations of this loss function. One with the static conflict matrix and without regularization, named \textit{ConsistencyLoss}. It has defined the conflicting emotions based on the models of affective psychology~\cite{plutchik1980general,russell1980circumplex}, and does not change the weights during learning. Another one, named \textit{GuidedConsistencyLoss}, utilizes the conflicting weight matrix as priors and learns the conflict matrix during training, but does not use any regularization. The last one, named \textit{RegularizedConsistencyLoss} learns the conflict matrix and also applies L1 regularization to diminish the sparsity of the learned weights.

\section{Experiments}
\label{sec:exp}
We conducted two main types of experiments. We trained the network model with a variety of loss functions on our new annotation data and did this for two types of annotations, one with the 7 universal emotions plus neutral, and the other for 14 emotional states. The second series of experiments are intended to show that our solution can be adjusted to control the granularity of the emotion description. The set of 14 emotions are the result of the fusion process described in Sect.~\ref{subsec:term-select} using $t=0.5$.

\subsection{Training Details and Evaluation Metrics}
We trained each model for 20 epochs using Adam-W with a learning rate of $1 \times 10^{-4}$. We used the extended B-AffectNet for training (Sect.~\ref{subsec:data_balance}), while the results are computed on the original AffectNet~\cite{mollahosseini2017affectnet} validation set.

Differently from the standard protocol where performance is validated through the prediction accuracy of the estimated vs ground-truth emotion label, here we aim at predicting the complete probability distribution over the emotions. Given that the ground-truth annotation is a distribution as well, it is required to adapt the evaluation metrics. Thus, in both the 8 and 14 emotion cases, we employ the following metrics: first, we use two distribution-based distances, namely the \textit{Jensen-Shannon} (JS) and \textit{Kullback–Leibler} (KL) divergence. We also evaluate the Cosine Similarity (CS) between the distribution vectors and the Pearson correlation coefficient. 

Nonetheless, we also compute Accuracy, Precision, Recall and F1-score. These are computed for the 8 emotions only as we only have ground-truth labels for that setting. To compute them, we get the ``dominant'' emotion from the predicted distributions as the one with highest probability and use it as predicted label. These are compared against the original emotion labels of AffectNet. This was done so to compare against current state-of-the-art models for emotion recognition in the standard sense.

\subsection{Model Performance on 8 Emotions}
The results for the experiments with 7 emotion classes plus neutral are shown in Table~\ref{tab:loss_comparison}. The ``ConsistencyLoss'' configuration shows the best overall performance. This evidences that 
for this specific scenario, a fixed prior knowledge encoding the semantics of emotions is relevant, and difficult to learn.
Nevertheless, the proposed loss that uses the semantics of conflicting emotions significantly improves upon the baseline using only Focal Loss in all metrics. Despite reporting results in terms of accuracy that are lower with respect to the current state-of-the-art, something expected given that we train on different annotations and for a different task, the relative improvement evidences that the such knowledge can be exploited to better capture the factors that influence the facial expression. 

In the experiments marked with LD (Learned Distribution), instead of using the distributions parameters from~\cite{RUSSELL1977273}, we have calculated them from the samples int the training data. This was done to assess the reliability of the definitions in~\cite{RUSSELL1977273}. Results show that the model trained on annotations generated using~\cite{RUSSELL1977273} overall perform better, supporting their reliability and suggesting a greater degree of generalization. 

Overall, results evidence that loss functions that penalize conflicting emotions excel in learning distributions for emotional states. It can help to describe more complex emotional representations not necessarily utilizing the most likely or dominant, but allowing a combination of different emotions given their probability.

\begin{table}[ht]
\centering
\caption{Comparison between predicted and ground-truth distributions for 14 emotions. Best results in bold.}
\label{tab:distribution_metrics}
\begin{adjustbox}{width=\linewidth}
\begin{tabular}{lccccc}
\toprule
\textbf{Experiment} &\textbf{KL ($\downarrow$)}& \textbf{JS ($\downarrow$)}& \textbf{Cosine ($\uparrow$)}&  \textbf{Pearson ($\uparrow$)}\\
\midrule
RegularizedConsistencyLoss & \textbf{2.50} & \textbf{0.24} & 0.58  & \textbf{0.42} \\
ConsistencyLoss            & 2.82 & 0.26 & \textbf{0.59} & 0.38 \\
\bottomrule
\end{tabular}
\end{adjustbox}
\end{table}

\begin{figure*}
    \centering
    \includegraphics[width=0.9\linewidth]{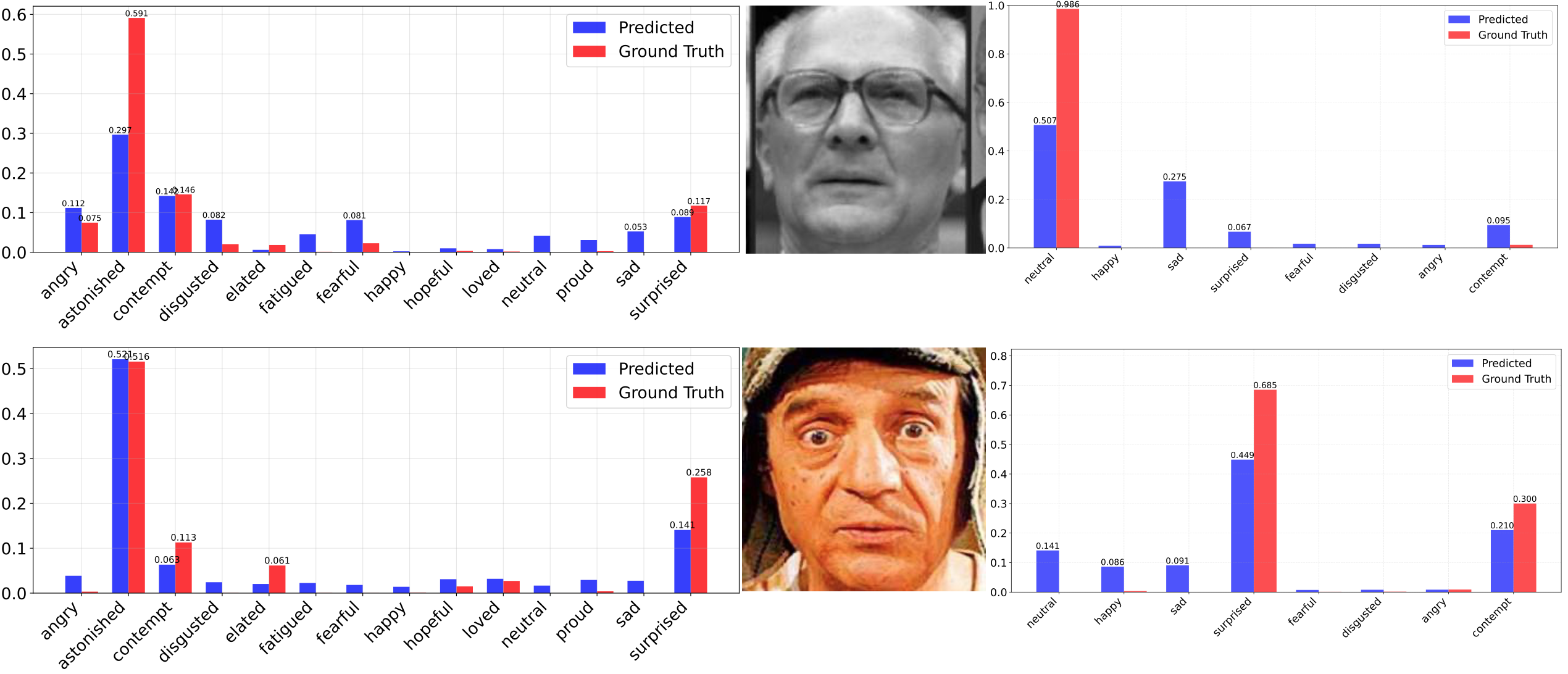}
    \caption{Comparison between the predicted and ground-truth distributions (both 14 and 8 emotions) for two example faces. The original images were labeled as ``neutral'' (top row) and ``surprise'' (bottom row) but the reader can likely see that the actual emotional state is more complicated, with subtle and combined emotions contributing to the overall facial expression. Our annotation and prediction model provides a preliminary solution for better capturing such nuisances.}
    \label{fig:ex_predictions}
\end{figure*}

\subsection{Model Performance on 14 Emotions}
We have chosen the best performing experiments from the universal emotions and utilized them to evaluate the performance in a scenario with 14 different emotional states. These were obtained through the fusion step described in Section~\ref{subsec:term-select} and, from 151 emotions, we get 13 (happy, contempt, elated, hopeful, surprised, proud, loved, angry, astonished, disgusted, fearful, sad, and fatigued) plus neutral. The results are shown in Table~\ref{tab:distribution_metrics}. It turns out that the regularized version of the consistency loss has slightly better performance than the basic one in this case, in contrast with the 8 emotions setting. A reason is that using a fixed prior for conflicting emotions when their description granularity increases, constrain the learning excessively. This suggests that adapting the conflicting schema can be beneficial in contexts where the set of emotions is larger and finer-grained. 


 
\smallskip
In both the 8 and 14 emotion cases, the performance of our baseline model look promising. The distribution-based metrics report quite low values (0.21/1.99 and 0.24/2.50) indicating that the predictions follow the real distributions rather accurately. Two examples are shown in Figure~\ref{fig:ex_predictions}. The analysis of the re-labeling process conducted with the user study (Sect.~\ref{subsec:data_val_res}) reports similar values, supporting the effectiveness claim and the proposed solution. In terms of accuracy instead, results are quite far from the state-of-the-art, yet we remark that a direct comparison is not possible as the protocol is different, the task simpler and the annotations as well have some degree of discrepancy (Sect.~\ref{sec:annot_val}). However, these test served as reference, as the objective would be gradually substituting the standard protocol with the one proposed in this paper, in the attempt of pushing forward research in the field of affective computing.

\section{Conclusions and Future Works}
In this paper, we proposed a novel solution to  describe and predict emotional states as probability distributions of emotions rather than a fixed one. Our re-labeling process allows us to better describe an emotional state by putting together different emotions and construct a macro state based on the joint probabilities. The results bring evidence that the relabeling is aligned with the emotions perceived on the images, and the user study corroborates the validity of the probabilistic labels. 
Even trough we generated the probability annotations, the distribution are not from the dataset itself but from a work on the literature that evaluates the problem of representation of emotions and how it is perceived by humans. We apply those distributions values to better expand the granularity of emotion representation, and to be faithful to a more generalist approach to facial expression recognition. We strongly believe that this is not circular validation since the distributions are not calculated from the data itself. Ultimately, the proposed approach represents a step forward towards a better modeling of emotions in automatic recognition. We also proposed a baseline network architecture implementing a novel loss function that uses some prior knowledge to ensure the semantic coherence. This allows the model to capture relations from the different emotional aspects, but without mixing conflicting emotional aspects. Such a prediction model could be used in future works in combination with language models in order to convert the probabilistic description into natural language.

On the other hand, there are still limitations and aspects to investigate further. For example, while it is true that emotional states are a combination of multiple subtle emotions, there is often one that dominates, and such an aspect should be taken into account. In addition, our re-labeling process required annotated data, which is still a burdensome and difficult problem that requires human intervention given that emotions are something purely human. These are all promising aspects to investigate in future research.



\section*{ETHICAL IMPACT STATEMENT}
This work explores a new approach for automatic emotion description and recognition from facial data. While such methods may support applications in human–computer interaction and affect-aware systems, they also raise ethical concerns. Facial expressions are not a reliable or universal proxy for internal emotional states, and predictions may be inaccurate or context-dependent. On the other hand, excessively accurate models could lead users to believe they are interacting with another person, raising potential safety concerns. In addition, facial emotion recognition involves sensitive personal data and may infringe on privacy if used without informed consent. 
All the data used in this paper comes from research datasets that are publicly available and have been obtained by signing the related agreements.

{\small
\bibliographystyle{ieee}
\bibliography{egbib}
}

\end{document}